\documentclass[conference]{IEEEtran}
\IEEEoverridecommandlockouts
\usepackage{cite}
\usepackage{amsmath,amssymb,amsfonts}
\usepackage{algorithmic}
\usepackage{graphicx}
\usepackage{textcomp}
\usepackage{xcolor}
\usepackage{adjustbox}
\usepackage{multirow}
\usepackage{tabularx}
\usepackage{hyperref}

\def\BibTeX{{\rm B\kern-.05em{\sc i\kern-.025em b}\kern-.08em
    T\kern-.1667em\lower.7ex\hbox{E}\kern-.125emX}}
\begin{document}

\title{Improving Graph Convolutional Networks with Transformer Layer in social-based items recommendation\
}


\makeatletter
\newcommand{\linebreakand}{%
  \end{@IEEEauthorhalign}
  \hfill\mbox{}\par
  \mbox{}\hfill\begin{@IEEEauthorhalign}
}
\makeatother

\author{
\IEEEauthorblockN{Thi Linh Hoang}
\IEEEauthorblockA{\textit{HMI Lab} \\
\textit{VNU University of Engineering and Technology}\\
Hanoi, Vietnam}\\[0.1cm]
\and
\IEEEauthorblockN{Tuan Dung Pham}
\IEEEauthorblockA{\textit{HMI Lab} \\
\textit{VNU University of Engineering and Technology}\\
Hanoi, Vietnam}\\[0.1cm]
\linebreakand
\IEEEauthorblockN{Viet Cuong Ta}
\IEEEauthorblockA{\textit{HMI Lab} \\
\textit{VNU University of Engineering and Technology}\\
Hanoi, Vietnam}\\[0.1cm]
}

\maketitle

\begin{abstract}
With the emergence of online social networks, social-based items recommendation has become a popular research direction.
Recently, Graph Convolutional Networks have shown promising results by modeling the information diffusion process in graphs.
It provides a unified framework for graph embedding that can leverage both the social graph structure and node features information.
In this paper, we improve the embedding output of the graph-based convolution layer by adding a number of transformer layers. 
The transformer layers with attention architecture help discover frequent patterns in the embedding space which increase the predictive power of the model in the downstream tasks.
Our approach is tested on two social-based items recommendation dataset, Ciao and Epinions and our model outperforms other graph-based recommendation baselines.

\end{abstract}


\section{Introduction}
With the explosive growth of online information, recommendation systems have played an important role in supporting users' decisions.
Due to the wide range of applications related to the recommendation system, the number of research works in this area is increasing fast.
An effective recommendation system must accurately capture user preferences and recommend items that users are likely to be interested in, which can improve user satisfaction with the platform and the user retention rate. In the context of an e-commerce and social media platform, both individual users' preferences and users' social relations are two information sources for choosing which items are most preferred by the user.

The general recommendation assumes that the users have static preferences and models the compatibility between users and items based on either implicit (clicks) or explicit (ratings) feedback and ignores social relations. The system predicts the user’s rating for the target item, i.e., rating prediction or recommends top-K items the user could be fascinated by, i.e., top-N recommendations. Most studies consider user-item interactions in the form of a matrix and take the recommendation as a matrix completion task. Matrix factorization (MF) is one of the most traditional collaborative filtering methods. MF learns user/item latent vectors to reconstruct the interaction matrix. Recent years have witnessed great developments in deep neural network techniques for graph data. Deep neural network architectures known as Graph Neural Networks (GNNs) \cite{b9} have been proposed to learn meaningful representations for graph data. GNN's main idea is to iteratively aggregate feature information from local graph neighborhoods using neural networks. The motivation for applying graph neural network methods to recommendation systems lies in two facets \cite{b2}: The majority of data in the Recommender System have a graph structure fundamentally and GNN algorithms are excellent for recording connections between nodes and graph data representation learning. \\

Graph Convolutional Network (GCN) \cite{b1} is a family of GNN models that could be used to distill graph-based information.
Therefore, in the context of social-based recommendation systems, the GCN can work on both user-item relations and user-user relations.
The main operation of the GCN is the graph convolution operation which could be considered as a more-generalized form of the standard image-based convolution.
In GCN, the state of the node is the same as that of the convolution operation in image processing, and the features are pooled from neighbored nodes which are defined by the local graph structure.
On the other hand, the Transformers \cite{b7} are shown to be the most effective neural network architectures flowing attention mechanism and similar to GNN.
In this paper, we aim to improve the standard GCN by adding several transformer layers.
With the attention mechanism from transformer layers, it helps to improve the feature space toward the downstream task, which is the link prediction task for social-based user recommendation. 
The proposed structure is then tested on two standard social-based item recommendation datasets, Ciao and Epinions.
The experiments show that the added attention layers could reduce the prediction errors of the standard GCN significantly. We provide the code base of this paper on \url{https://github.com/linhthi/ts}.

Our paper is organized as follows: in Section II, the related works are presented; our proposed architecture is introduced in Section III; the experiments are given in Section IV; and Section V is the conclusion.

\section{Related work}
Traditional recommendation using matrix factorization (MF) techniques: Probabilistic matrix factorization (PMF) \cite{b3}  takes a probabilistic approach in solving the MF problem  $M  \approx UV_{T}$. Neural Collaborative Filtering (NeuMF) \cite{b4} extends the MF approach by passing the latent user and item features through a feed-forward neural network. \par

Most individuals have online social connections. Relationships between users and their friends are varied and they are usually related to each other. Based on this assumption, the user's preferences may be similar to or influenced by their related peers. They also tend to recommend similar items. Thus, a social-based item recommender system was introduced to extract information about user’s preferences from their social relationship. With the success of GNN in molecular biology with small networks as protein interaction, many researchers try to apply GNN on large-scale graphs like social networks.\par

One of the most GNN fundamental approaches for node embedding is based on neighborhood normalization. In images, a convolution is computed with the weighted sum of the neighbor’s features and weight-sharing, thanks to the neighbor's relative positions. With graph-structured data, the convolution process is different. Graph Convolutional Networks (GCN) \cite{b1} are the most popular of neighborhood approaches. Convolution methods in graph convolutional Convolutional networks can be divided into two categories: Spectral convolution \cite{b10} and Spatial convolution \cite{b8}.

Spectral-based convolution filters are inherited from signal processing techniques. Spectral GCN \cite{b10}, i.e. the filter of the convolutional network and the image signal are transferred to the Fourier domain for processing at the same time. Spectral GCN can be defined by a function of frequency. Therefore, the information on any frequency can be found using this method. However, All of spectral GCN methods rely on Laplacian matrices, which must operate on the entire graph structure. The forward/inverse Fourier transformation of a graph could cost a lot of computation resources. Moreover, Spectral GCN makes a trained model difficult to apply to other problems since the filter resulting from one graph can not be applied to others.

The spatial GCN \cite{b8} for graphs belongs to the spatial domain convolution, that is, the nodes in the graph are connected in the spatial domain to achieve a hierarchical structure, and then convolution. In general, spatial convolutions in graphs require fewer computational resources and their transferability is better than spectral convolutions. The non-spectral method, i.e. the spatial domain GNN method needs to find a way to deal with different numbers of neighbors of each node. \\

Another research direction of GNN that applying attention mechanisms. Attention mechanism originates from the field of Natural Language Processing in tasks such as machine translation. Recently when attention is applied to graphs also gives quite good results compared to other methods \cite{b2} \cite{b5} \cite{b6} \cite{b11} \cite{b12} \cite{b13}. Graph Attention Network (GAT) \cite{b5} expands the basic aggregation function of the GCN layer, assigning different importance to each edge through the attention coefficients. GraphRec \cite{b2} using attention network on social information and user opinions. Graph Transformer \cite{b6} developed after that uses Transformer - a more complex attention function, but it still has a problem with the difficulty of positional encoding on graph data.

\section{Methodology}
\subsection{Definition}
We describe a recommendation system as an directed $\mathcal{G}=(\mathrm{V}, \mathcal{E})$ where is size $N$ of set nodes $v_{i} \in \mathbb{V}$ and edges $\left(v_{i}, v_{j}\right) \in \mathcal{E}$. The node features are denoted as $X=\left\{x_{1}, \cdots, x_{N}\right\} \in \mathbb{R}^{N \times C}$, and the adjacency
matrix is defined as $A \in \mathbb{R}^{N \times N}$ which associates each edge $\left(v_{i}, v_{j}\right)$ with its element $A_{i j} .$ The node degrees are given by $d=\left\{d_{1}, \cdots, d_{N}\right\}$ where $d_{i}$ computes the sum of edge weights connected to node $i$. We define $D$ as the degree matrix whose diagonal elements are obtained from $d$. The graph $\mathcal{G}$ represents data in the recommender system with nodes and edges information is input. For simplicity, the homogeneous graph is used instead of the heterogeneous graph. 
\\


\subsection{Graph Convolutional Network}
Graph Convolution Network (GCN) is originally developed by Kipf \& Welling (2017). The feed-forward propagation in GCN is recursively conducted as

\begin{equation}
H^{(l+1)}=\sigma\left(\hat{A} H^{(l)}W^{(l)}\right)
\end{equation}

\noindent
where $H^{(l+1)} = \left\{h_{1}^{(l+1)}, \cdots, h_{N}^{(l+1)}\right\}$ are the hidden vectors of the $l$ -th layer with $h_{i}^{(l)}$ as the hidden feature for node $i ; \hat{A}=\hat{D}^{-1 / 2}(A+I) \hat{D}^{-1 / 2}$ is the re-normalization of the adjacency matrix, and $\hat{D}$ is the corresponding degree matrix of $A+I$ ; $\sigma(\cdot)$ is a nonlinear function, i.e. the ReLu function; and $W^{(l)} \in \mathbb{R}^{C_{l} \times C_{l-1}}$ is the filter matrix in the $l$ -th layer with $C_{l}$ refers to the size of $l$ -th hidden layer. We denote a layer GCN computed by Equation 1 as a Graph Convolutional layer (GC layer) in what follows.

\subsection{Transformer Encoder}
The Transformer \cite{b7} is the first transduction model relying entirely on self-attention to compute representations of its input and output without using sequence-aligned RNNs or convolution. Following relevant research \cite{b6} \cite{b11} \cite{b12} \cite{b13}, the graph can be the input of a Transformer instead of traditional data input - sequences, thus we use Transformer-like a component of network embedding module. Firstly, we update the hidden feature $h$  of the $i$ \'th node in a graph  from layer $l$  to layer $l-1$  as follows:
\begin{equation}
h_{i}^{\ell+1}=\text { Attention }\left(Q^{\ell} h_{i}^{\ell}, K^{\ell} h_{j}^{\ell}, V^{\ell} h_{j}^{\ell}\right)
\end{equation}

i,e.,
\begin{equation}
h_{i}^{\ell+1}=\sum_{j \in \mathcal{N}(i)} w_{i j}\left(V^{\ell} h_{j}^{\ell}\right)
\end{equation}

\begin{equation}
\text { where } w_{i j}=\operatorname{softmax}_{j}\left(Q^{\ell} h_{i}^{\ell} \cdot K^{\ell} h_{j}^{\ell}\right)
\end{equation}
where $j \in \mathcal{N}(i)$ denotes the set of neighbor nodes of node $i$ in graph and $Q^{\ell}, K^{\ell}, V^{\ell}$ are learnable linear weights (denoting the Query, Key and Value for the attention computation, respectively). The attention mechanism is performed parallelly for each node in the neighbor nodes to obtain their updated features in one shot—another plus point for Transformers over RNNs, which update features node-by-node.\par

Multi-head Attention: Getting this straightforward dot-product attention mechanism to work proves to be tricky. Bad random initializations of the learnable weights can destabilize the training process. We can overcome this by parallelly performing multiple 'heads' of attention and concatenating the result (with each head now having separate learnable weights):
\begin{equation}
h_{i}^{\ell+1}=\text { Concat }\left(\text { head }_{1}, \ldots, \text { head }_{K}\right) O^{\ell}
\end{equation}
\begin{equation}
\text { head }_{k}=\text { Attention }\left(Q^{k, \ell} h_{i}^{\ell}, K^{k, \ell} h_{j}^{\ell}, V^{k, \ell} h_{j}^{\ell}\right)
\end{equation}
where $Q^{k, \ell}, K^{k, \ell}, V^{k, \ell}$ are the learnable weights of the $k$ 'th attention head and $O^{\ell}$ is a down-projection to match the dimensions of $h_{i}^{\ell+1}$ and $h_{i}^{\ell}$ across layers.\par


Transformers overcome the issue of the individual feature/vector entries level, concatenating across multiple attention heads each of which might output values at different scales can lead to the entries of the final vector $h_{i}^{\ell+1}$ having a wide range of values with LayerNorm, which normalizes and learns an affine transformation at the feature level. Additionally, scaling the dot-product attention by the square root of the feature dimension helps counteract the issue that the features for nodes after the attention mechanism might be at different scales or magnitudes.
Finally, the authors propose another 'trick' to control the scale issue: a position-wise 2-layer MLP with a special structure. After the multi-head attention, they project $h_{i}^{\ell+1}$ to a (absurdly) higher dimension by a learnable weight, where it undergoes the ReLU non-linearity and then projected back to its original dimension followed by another normalization:
\begin{equation}
 h_{i}^{\ell+1}=\mathrm{LN}\left(\mathrm{MLP}\left(\mathrm{LN}\left(h_{i}^{\ell+1}\right)\right)\right)   
\end{equation}

\subsection{Graph Transformer Network}
The architecture of the proposed method for network embedding the social graph and user-item interaction graph is shown in Figure 1, the model still follows the GNN to learn node embedding from graph data, but with modification toward a more general solution. The model uses both Graph Convolution layers and transformer layers as an encoder approach. The transformer layer is used because it is similar to GNN models that use the attention mechanism. The difference is that the Transformer uses a more complex attention function.
\par

\begin{figure*}[!htbp]
    \centering
    \includegraphics[width=1\textwidth]{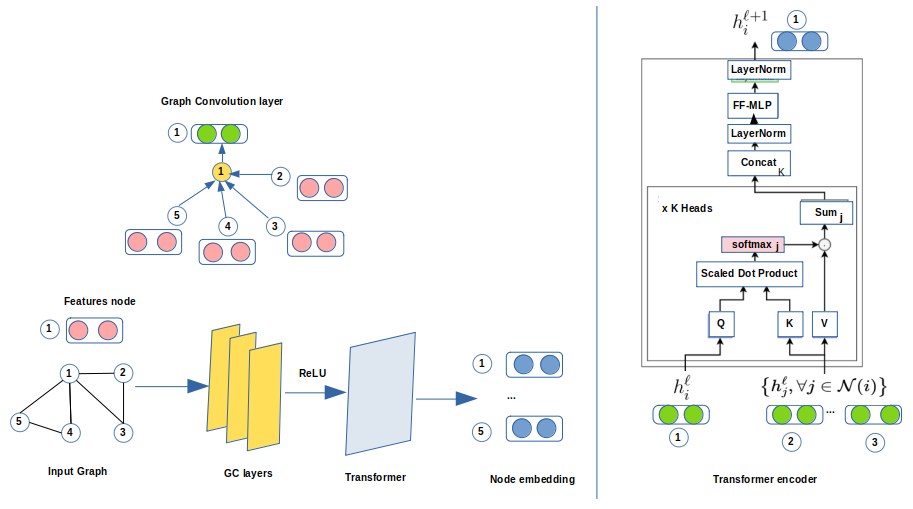}
    \caption{Illustration of the architecture of the proposed model}
\end{figure*} 

In general, We use two graph convolution layers to embed nodes from the graph, with other experiments we will discuss in section IV. As we mentioned in the previous section, the graph convolution layer follows the local aggregation.
The input of graph convolution layer 1 is the normalization adjacency matrix (for simple we define adjacency matrix $A$ size $(N \times N)$ where N is the number of nodes needed to embed) and matrix $X$ of nodes features size $(N \times F)$. Output is matrix embedded $H^{(1)}$ with size $(N, hidden\_size)$.\par

\begin{equation}
    H^{(0)} = X
\end{equation}
\begin{equation}
    H^{(1)} = ReLU(AH^{(0)}W^{(0)})
\end{equation}
\begin{equation}
    H^{(2)} = ReLU(AH^{(1)}W^{(1)})
\end{equation}

After the step in graph convolution layer 2, we have new features of nodes in the graph. Thus, we feed the new graph to the transformer layer encoder to improve embedding nodes with an attention mechanism.
\begin{equation}
    H^{(3)} = transformer\_encoder(H^{(2)})
\end{equation}

When the nodes are embedded by a transformer encoder, the pairs of nodes in the graph are selected to predict the score between them. Assuming, to calculate the score/relation of node i and node j. At first, we need to combine them to feed the function (e.g. concat, dot product). In this case, we use concatenation.
\begin{equation}
    H(i,j) = H(i)\mathbin\Vert H(j)
\end{equation}

Finally, $H(i,j)$ is used to predict the result, for the simple Linear layer to complete this task.
\begin{equation}
    \hat{y}_{ij} = W*H(i,j) + b
\end{equation}
where W is the weight of the Linear layer and b is bias.

The output of the transformer layer is an embedding of each relationship in the graph, between a user \textit{i} and an item \textit{j}
To train the model parameters effectively, the output embedding is connected with downstream tasks.
In the context of item recommendations, one could add a linear layer with regression output.
The output is the rating prediction which specifies how preferred the users could choose/select the items.
The loss in this case is the standard mean squared error between the prediction outputs and the targets.

\begin{equation}
   Loss = \frac{1}{N} \sum(\hat{y}_{ij} - y_{ij})^2
\end{equation}

where N is the number of ratings, and $y_{ij}$ is the ground trust rating assigned by the user i on the item j.

\section{Experiments and Results}
\subsection{Dataset}

In this paper, Ciao and Epinions \footnote{http://www.cse.msu.edu/~tangjili/index.html} are chosen datasets to evaluate the performance of the model, which are taken from popular social networking websites Ciao and Epinions. At Epinions and Ciao, visitors can read reviews about a variety of items to help them decide on a purchase or they can join for free and begin writing reviews that may earn them rewards and recognition. To post a review, members need to rate the product or service on a rating scale from 1 to 5 stars. Every member of Epinions maintains a “trust” list which presents a network of trust relationships between users, and a “block (distrust)” list which presents a network of distrust relationships. This network is called the “Web of trust”, and is used by Epinions and Ciao to reorder the product reviews such that a user first sees reviews by users that they trust.\\



 Other statistics of the two datasets are respectively presented in Table I. Each dataset contains two files, one is the rating data of the item given by the user, and another stores the trust network data.
\begin{table}[!htbp]
    \centering
    \caption{The datasets statistics }
    \begin{tabular}{ |c|c|c| }

        \hline
         \textbf{Dataset} & \textbf{Ciao} & \textbf{Epinions}\\
        \hline
         number of users & 7375 &18088 \\
        \hline
         number of items & 105114 & 261649    \\
        \hline
        number of ratings & 288319 &764352 \\
        \hline
        density (ratings) & 0.0372\% & 0.0161\% \\
        \hline
        number of social connections & 111781 & 355813 \\
        \hline
        density (social connections) & 0.2055\% & 0.1087\% \\
        \hline
        mean (ratings) & 4.16 & 3.97 \\
        \hline
    \end{tabular}
\end{table}

\subsection{Evaluation metrics}
Other recommendation systems regularly use the Hit ratio or Top@K ranking to evaluate metrics, but in this work, we consider this problem as a regression problem. The prediction quality of our proposed approach in comparison with other collaborative filtering and trust-aware recommendation methods, we use two standard metrics, the Mean Absolute Error (MAE) and the Root Mean Square Error (RMSE).\\

The metric MAE is defined as:  
\begin{equation}
M A E=\frac{1}{n} \sum_{(u, i) \in T}\left|\hat{y}_{u i}-y_{u i}\right|
\end{equation}

The metric RMSE is defined as: 
\begin{equation}
R M S E=\sqrt{\frac{1}{n} \sum_{(u, i) \in T}\left(\hat{y}_{u i}-y_{u i}\right)^{2}}
\end{equation}

where $y_{ij}$ denotes the rating user $i$ gave to item $j$, $\hat{y}_{ij}$ denotes the rating user $i$ gave to item $j$ as predicted by method, and $n$ denotes the number of tested ratings set $T$.\par

From the definitions, we can see that smaller MAE or RMSE values mean better performance.

\subsection{Training specifications}
For each dataset, we use 60\% as a training set to learn parameters, 20\% as a validation set to tune hyper-parameters, and 20\% as a test set for the final performance comparison. We chose Adam as the optimizer to train the network since Adam shows much faster convergence than standard stochastic gradient (SGD) with momentum in this task. For the hidden dimension size d, we tested the value of [8, 16, 32, 64, 128]. The batch size and learning rate were searched in [32, 64, 128, 512] and [0.005, 0.001, 0.05, 0.01]. For the number of Graph Convolution layers in GCN and GTN we tested the value of [1, 2, 3]. And to find the affection of multi-head in the Transformer, we will test on [1, 2, 3] head. All weights in the newly added layers are initialized with a Gaussian distribution.\par
The networks are trained for 50 epochs and make use of early stopping to avoid overfitting. The time to train each model takes about two hours.\\

We use Google Colab with a P100 GPU for training. During the training, the adjacency list and the feature matrix of nodes are placed in CPU memory due to their large size. However, during the convolution step, each GPU process needs access to the neighborhood and feature information of nodes in the neighborhood. Accessing the data in CPU memory from GPU is not efficient. To solve this problem, we use a re-indexing technique to create a sub-graph $\hat{G} = (\hat{V}, \hat{E})$ containing nodes and their neighborhood, which will be involved in the computation of the current minibatch. A small feature matrix containing only node features which relevant to the computation of the current minibatch is also extracted such that the order is consistent with the index of nodes in $\hat{G}$.
The adjacency list of $\hat{G}$ and the small feature matrix is fed into the GPU at the start of each minibatch iteration, so that no communication between the GPU and CPU is needed during the convolve step, greatly improving GPU utilization. The training procedure has alternating usage of CPU and GPU. The model computations are in GPU, whereas extracting features, re-indexing, and negative sampling are computed on CPUs. \\

We train GCN first to find the better parameters. Then we apply the parameters in the GTN model. As we show in Figure 2, experiments training on GTN are quite similar to GCN, but GTN seems to have better performance.

\subsection{Results}
We test the two the standard Graph Convolution Network (GCN) \cite{b1} and our proposed model Graph Transformer Network (GTN) in both the Ciao and Epinions link prediction datasets.
Three other methods are used to compare with the proposed method including Probabilistic Matrix Factorization (PMF), Neural Collaborative Filtering (NeuMF), and Graph Neural Network for social recommendation (GraphRec).\\

\textbf{PMF} \cite{b3}:  an approach for link predictions based on the standard matrix decomposition. This model views the rating as a probabilistic graphical model. Given prior for the latent factors for users and items the equivalent problem is to minimize the square error.\\

\textbf{NeuMF}\cite{b4}: This method is state-of-the-art matrix factorization model with neural network architecture. The original implementation is for ranking tasks using implicit feedback and we adjust it to a regression problem for rating prediction. \\

\textbf{GraphRec} \cite{b2}: a Graph Neural Network framework to model graph data in social recommendation coherently for rating prediction. The model consists of three components: user modeling, item modeling, and rating prediction. The user modeling component is to learn the factors of users. As data in the social recommender system includes two different graphs, i.e., a trust graph and a user-item graph. The item modeling component is to learn the latent factors of items. The rating prediction component is to learn model parameters via prediction by integrating user and item modeling components.\\

With the default parameters, the training loss of the baseline model is described in Figure 2. The loss of GraphRec and NeuMF at the few first step has been significantly reduced while GTN and GCN have a quite small loss. Loss of PMF is still decreasing at the 50th epoch.

\begin{figure}[h]
    \centering
    \includegraphics[width=0.5\textwidth]{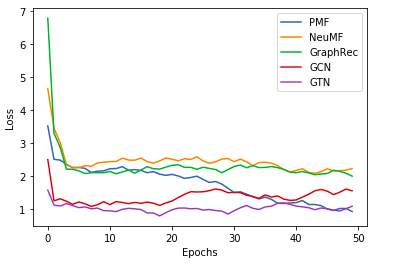}
    \caption{Loss in the training process with our model and baseline models on Ciao dataset}
\end{figure}

\begin{table}[!htbp]
    \centering
    \caption{Performance comparison of different recommender systems}
    \begin{tabular}{ |c|c|c|c|c|c|c| }
        \hline
         \multirow{2}{*}[-2pt]{\textbf{Dataset}} & \multirow{2}{*}[-2pt]{\textbf{Metric}} & \multirow{2}{*}[-2pt]{\textbf{PMF}} & \multirow{2}{*}[-2pt]{\textbf{NeuMF}} & \textbf{Graph} & \multirow{2}{*}[-2pt]{\textbf{GCN}} & \multirow{2}{*}[-2pt]{\textbf{GTN}}\\
         
         ~ & ~ & ~ & ~ & \textbf{Rec} & ~ & ~ \\
        \hline
         \textbf{Ciao} & MAE & 0.8184 & 0.8052 & 0.7834 & 0.8270 & 0.7641 \\
        \hline
         \textbf{Ciao} & RMSE & 1.0581 & 1.0439  & 1.0090 & 1.0605 & 0.9732     \\
        \hline
        \textbf{Epinions} & MAE & 0.9713 & 0.9072 & 0.8524 & 0.8956 & 0.8436 \\
        \hline
        \textbf{Epinions} & RMSE & 1.1829 & 1.1411 & 1.1078 & 1.1680 & 1.0139  \\
        \hline
    \end{tabular}
    
\end{table}

Table II shows the performance of the two models and baselines on the Ciao and Epinions dataset.
NeuMF obtains much better performance than PMF. Both methods only utilize the rating information. GNN-based methods, e.g., GraphRec, GCN, and GTN perform better than matrix factorization methods. Furthermore, GraphRec usually performs better than GCN. Because GraphRec uses full graph information to learn factors, while GCN just uses neighbors in the training set. Our GTN model achieves the best performance compared to all other baselines on all the datasets. It demonstrates that the GTN can learn node embedding more effectively for social data.

\textbf{The experimental results with the number of Graph Convolution layers}
The depth of neighbor relations affects the aggregate neighborhood information process. As we mentioned in the training specifications section, we test on 1, 2, and 3 Graph Convolution layers to find the best model. Following on the result is shown in Table III - training on Epinions dataset and RMSE metric, with 2 Graph Convolution layers, models give the best result.

\begin{table}[!htbp]
    \centering
    \caption{Performance comparison of different size of graph convolution layers on Epinions dataset with RMSE}
    \begin{tabular}{ |c|c|c|c| }

        \hline
         \textbf{Model} & \textbf{1 GC} & \textbf{2 GC} & \textbf{3 GC}\\
        \hline
         \textbf{GCN} & 1.1608 & 1.0456 & 1.0978\\
        \hline
         \textbf{GTN} & 1.0139 & 0.9743 & 1.0034 \\
        \hline
    \end{tabular}
\end{table}

\textbf{The experimental results with the size of multi-head attention in Transformer layer}
In practice, given the same set of queries, keys, and values we may want our model to combine knowledge from different behaviors of the same attention mechanism. Thus, it may be beneficial to allow our attention mechanism to jointly use different representation subspaces of queries, keys, and values.

To this end, instead of performing a single attention pooling, queries, keys, and values can be transformed with $h$  independently learned linear projections. 
The experiment of the number of muti-head attention is described in Table IV. As the result, with 3 attention heads, the model gives the best performance.

\begin{table}[!htbp]
    \centering
    \caption{Performance comparison of different size of multi-head attention on Epinions dataset}
    \begin{tabular}{ |c|c|c|c| }

        \hline
         \textbf{Metric} & \textbf{1 head} & \textbf{2 heads} & \textbf{3 heads}\\
        \hline
         \textbf{MAE} & 0.8439 & 0.8327 & 0.8123\\
        \hline
         \textbf{RMSE} & 1.0139 & 0.9957 & 0.9841 \\
        \hline
    \end{tabular}
\end{table}

\textbf{Time and Memory usage}: As shown in Table V, the average training time and the maximum memory usage on each model are acceptable. However, GTN proved to be at a disadvantage, which has memory usage and time training higher than other models. To remedy this situation, We also did some experiments to reduce both time training and memory usage.
\begin{table}[!htbp]
    \centering
    \caption{The comparison of time training and memory usage}
    \begin{tabular}{ |c|c|c|c|c|}

        \hline
         \textbf{Model} & \textbf{Dataset} & \textbf{Time training (h)} & \textbf{CPU (gb)} & \textbf{GPU (gb)}\\
        \hline
        \textbf{PMF} & \textbf{Ciao} & 0.45 & 2 & 0  \\
        \hline
        \textbf{NeuMF} & \textbf{Ciao} & 1.2 & 2 & 4.5     \\
        \hline
        \textbf{GraphRec} & \textbf{Ciao} & 2.1 & 2.35 & 5.5     \\
        \hline
        \textbf{GCN} & \textbf{Ciao} & 1.6 & 2.47 & 7.8     \\
        \hline
        \textbf{GTN} & \textbf{Ciao} & 2.05 & 2.93  & 11.4     \\
        \hline
    \end{tabular}
\end{table}

\section{Conclusion}
In this work, we have proposed an approach for improving the GCN for predicting ratings in social networks.
Our model is expanded from the standard model with several layers of transformer architecture.
The main focus of the paper is on the encoder architecture for node embedding in the network.
Using the embedding layer from the graph-based convolution layer, the attention mechanism could rearrange the feature space to get a more efficient embedding for the downstream task.
The experiments showed that our proposed architecture achieves better performance than GCN on the traditional link prediction task.


\end{document}